\documentclass[10pt,twocolumn,letterpaper]{article}

\usepackage{wacv}
\usepackage{times}
\usepackage{epsfig}
\usepackage{graphicx}
\usepackage{amsmath}
\usepackage{amssymb}
\usepackage{dblfloatfix}
\usepackage{flushend}

\usepackage[dvipsnames]{xcolor}
\usepackage{algpseudocode,algorithm,algorithmicx}
\usepackage{subfiles}

\usepackage{comment}

\usepackage{cuted}
\usepackage{capt-of}

\newcommand*\Let[2]{\State #1 $\gets$ #2}
\algrenewcommand\algorithmicrequire{\textbf{Precondition:}}
\algrenewcommand\algorithmicensure{\textbf{Postcondition:}}
\def\wacvPaperID{1238} 

\wacvfinalcopy 

\ifwacvfinal
\def\assignedStartPage{1} 
\fi

\ifwacvfinal
\usepackage[breaklinks=true,bookmarks=false]{hyperref}
\else
\usepackage[pagebackref=true,breaklinks=true,colorlinks,bookmarks=false]{hyperref}
\fi

\usepackage{xcolor,colortbl}
\usepackage{booktabs}
\usepackage{multirow}

\ifwacvfinal
\else
\fi

\begin{document}
\title{\vspace{-1.2cm}DOT: Dynamic Object Tracking for Visual SLAM}

\author{
Irene Ballester$^{1,2}$\\
{\tt\small iballestercampos@gmail.com  }
\and
Alejandro Fontan$^{1,2}$\\
{\tt\small alejandro.fontanvillacampa@dlr.de}
\and
Javier Civera$^{1}$\\
{\tt\small jcivera@unizar.es}
\and
Klaus H. Strobl$^{2}$\\
{\tt\small klaus.strobl@dlr.de}
\and
Rudolph Triebel$^{2,3}$\\
{\tt\small rudolph.triebel@dlr.de}
\and
$^{1}$ University of Zaragoza
\and
$^{2}$ German Aerospace Center (DLR)\\
\and
$^{3}$ Technical University of Munich\\
}


\maketitle

\begin{strip}\vbox{\vspace{-1.0cm}
\includegraphics[width=\textwidth]{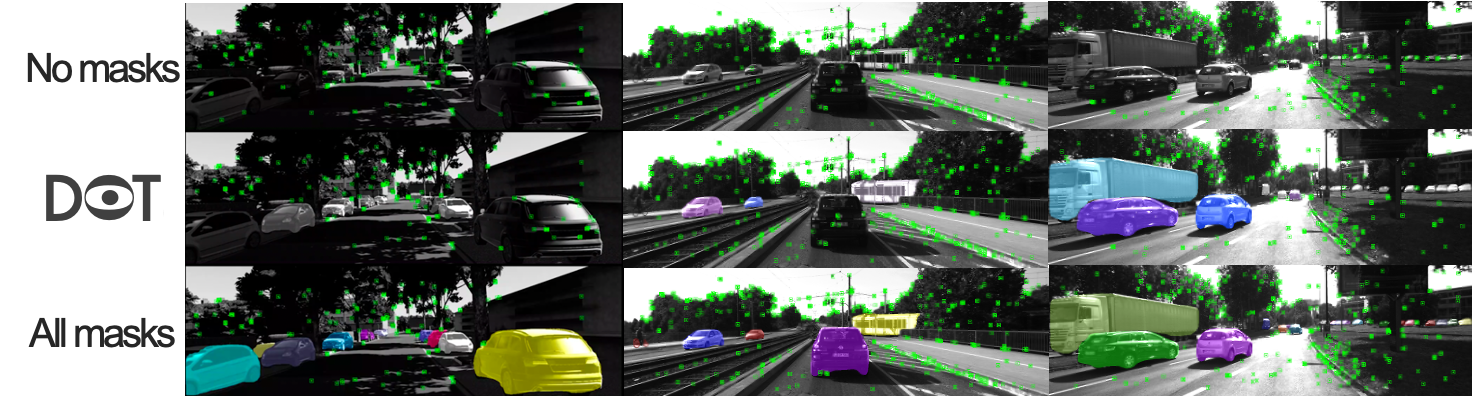}
\captionof{figure}{\textbf{Top row}: The frames correspond to ORB-SLAM2 \cite{Mur_Artal_2017} estimating the trajectory of the camera from the stream of images in The KITTI Benchmark \cite{KITTI}.  \textbf{Middle row}: Modified ORB-SLAM2 that works with the segmentation masks generated by DOT, which distinguish between moving and static objects. \textbf{Bottom row}: Modified ORB-SLAM2 using the segmentation masks provided by Detectron2 \cite{wu2019detectron2}, that encode all potential dynamic objects. Note how from the most static scene (left column) to the most dynamic one (right column), DOT is capable to avoid moving objects while keeping the static ones. DOT achieves a trade-off between those two opposing scenarios by estimating the actual motion state of the objects in order to get higher tracking robustness and accuracy.}
}
\label{fig:feature-graphic}
\end{strip}

\begin{abstract}
In this paper we present DOT (Dynamic Object Tracking), a front-end that added to existing SLAM systems can significantly improve their robustness and accuracy in highly dynamic environments. DOT combines instance segmentation and multi-view geometry to generate masks for dynamic objects in order to allow SLAM systems based on rigid scene models to avoid such image areas in their optimizations. 

To determine which objects are actually moving, DOT  segments first instances of potentially dynamic objects and then, with the estimated camera motion, tracks such objects by minimizing the photometric reprojection error. 
This short-term tracking improves the accuracy of the segmentation with respect to other approaches. In the end, only actually dynamic masks are generated. We have evaluated DOT with ORB-SLAM 2 \cite{Mur_Artal_2017} in three public datasets. Our results show that our approach improves significantly the accuracy and robustness of  ORB-SLAM 2, especially in highly dynamic scenes.
\end{abstract}


\section{Introduction}

Simultaneous Localization and Mapping, commonly known by its acronym SLAM, is one of the fundamental capabilities for the autonomous navigation of robotic platforms \cite{cadena2016past}. Its goal is the joint estimation of the robot motion and a map of its surroundings, from the information of its embedded sensors. Visual SLAM, for which the sensors are mainly, or exclusively, cameras, is one of the most challenging yet relevant configurations. 


Despite the significant advances in SLAM in the last two decades, most state-of-the-art systems still assume a static environment, where the relative position between the scene points does not change and the only motion is done by the camera. With this assumption, SLAM models attribute the visual changes exclusively to the relative camera motion. 
A usual approach \cite{mur2015orb,Mur_Artal_2017} is modeling dynamic areas as outliers, ignoring them during the pose tracking and map estimation processes. However, for several frames, until such dynamic areas are discarded as outliers, their data is used in the SLAM optimization, hence introducing errors and inconsistencies in the estimation of the map and the camera poses. Moreover, for feature-based SLAM methods, that track a small number of salient image points, the errors produced by a relatively small number of matches in dynamic areas are relevant and can lead to the system failure. 

The world and the real applications in which a robot or an AR system must operate is far from being static. We can cite as representative examples the autonomous navigation of cars or drones, AR in crowded scenes or even planetary exploration tasks, where the poor texture makes SLAM systems precarious in the presence of shadows or other robots. Developing SLAM systems that are sufficiently robust to operate in highly dynamic environments is then essential for many applications. 

As shown in the Figure \ref{fig:feature-graphic}, this work aims to develop an image processing strategy that improves the robustness of a visual SLAM system in dynamic environments. Our specific contribution is the development of ``Dynamic Object Tracking'' (DOT), a front-end that combines instance segmentation with multi-view geometry to track the camera motion, as well as the motion of the dynamic objects, using direct methods \cite{engel2017direct}. The result of this pre-processing is a mask containing the dynamic parts of each image, that a SLAM system can use to avoid making  correspondences in such regions. 

Our experimental results in three different public datasets show that our combination of semantic segmentation and geometry-guided tracking outperforms the state of the art in dynamic scenes. We also find relevant that DOT is implemented as an independent front-end module, and hence easy-to-plug in existing SLAM systems. As DOT includes short-term mask tracking, we avoid the segmentation of all frames in the sequence, with significant savings in computation. Finally, although we tuned and evaluated DOT for the specific domain of car navigation, our strategy would be valid for other applications.





\section{Related Work}
\label{sec:related_work}

SLAM in dynamic environments is an open research problem with a large scientific bibliography. We will divide the different approaches into three main categories.

The first category, and the most general one, models the scene as a set of non-rigid parts, hence including deformable and dynamic objects \cite{Newcombe_2015_CVPR,innmann2016volume,lamarca2019defslam}. While this research line is the most general, it is also the most challenging one. In this paper we will assume intra-object rigidity, which is the premise behind the other two categories of dynamic visual SLAM. 

The second category aims to improve the accuracy and robustness of visual SLAM by reconstructing only the static part of a scene. Dynamic objects are segmented out and ignored for camera pose tracking and map estimation. Along this line, DynaSLAM \cite{DBLP:journals/corr/abs-1806-05620}, built on top of ORB-SLAM2 \cite{Mur_Artal_2017}, aims to estimate a map of the static part of the scene and re-use it in long-term applications. Dynamic objects are removed by combining 1) semantic segmentation for potentially moving objects, and 2) multi-view geometry for detecting inconsistencies in the rigid model. Mask R-CNN \cite{MaskRCNN} is used for semantic segmentation, which detects and classifies the objects in the scene into different categories, some of which have been pre-set as potentially dynamic (\eg, car or person). DynaSLAM was designed to mask out all the potentially mobile objects in the scene, which results in a lower accuracy than the original ORB-SLAM2 in scenes containing potentially mobile objects that are not actually moving (\eg, scenes with many parked cars). The aim of this work is, precisely, to overcome this problem as only those objects that are moving at that precise moment will be labeled as dynamic.

Another work that has a similar approach is StaticFusion \cite{StaticFusion}, a dense RGB-D visual SLAM system where segmentation is performed by using the 3D reconstruction of the scene background as a way of propagating the temporal information about the static parts of the scene.

Finally, the third line of work in dynamic visual SLAM, which goes beyond the segmentation and suppression of dynamic objects, includes works such as MID-Fusion \cite{MidFusion}, MaskFusion \cite{MaskFusion}, DynSLAM \cite{Barsan_2018} and ClusterVO\cite{ClusterVO}. Their aim is to simultaneously estimate the poses of the camera and multiple dynamic objects. For that purpose, in  MID-Fusion \cite{MidFusion} and MaskFusion \cite{MaskFusion}  sub-maps of each possible moving object are created and a joint estimation of both the objects and camera poses is carried out.

Most of the systems mentioned \cite{MidFusion, Barsan_2018, MaskFusion,ClusterVO,DBLP:journals/corr/abs-1806-05620} involve deep learning methods, which in some cases cannot be currently implemented in real-time due to bottleneck imposed by the limited frequencies of the segmentation network. The contribution developed in this work eliminates the requirement to segment all the frames, which allows the system to be independent of the segmentation frequency of the network, thus enabling its implementation in real time.

\section{DOT}
\label{sec:dot}

\subsection{System Overview}

\begin{figure*}[]
\begin{center}
\includegraphics[width=0.95\textwidth]{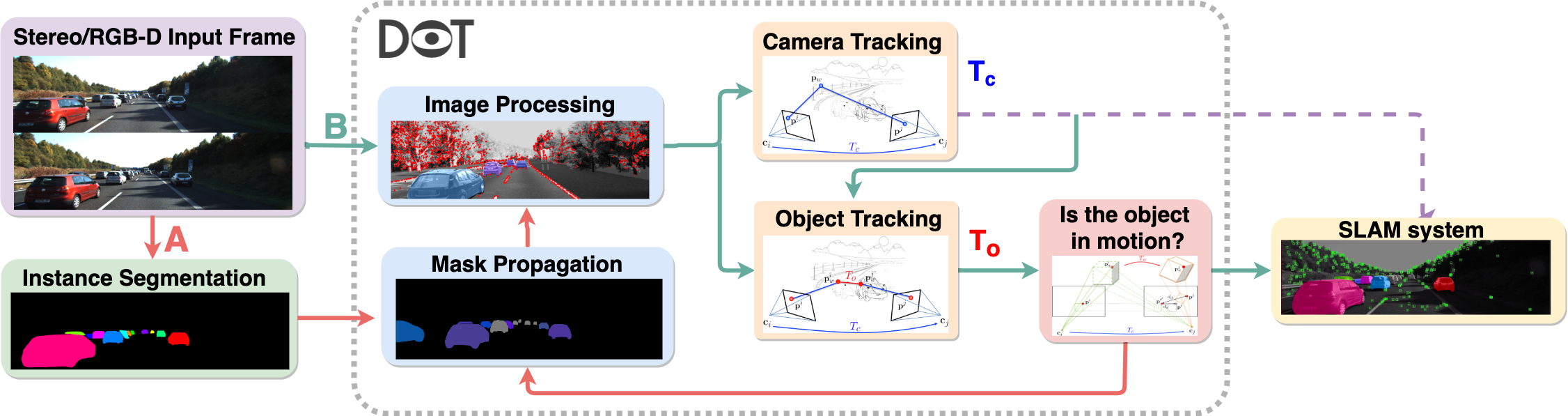}
\end{center}
  \caption{\textbf{Overview of DOT}. Path A (red), shows the processing for frames that get a segmentation mask from the network. Path B (green), shows the processing for frames that will acquire a segmentation mask geometrically propagated by DOT.}
    \label{fig:diagramaDOT}
\end{figure*}

Figure \ref{fig:diagramaDOT} shows an overview of our proposal.
The input to DOT are either RGB-D or stereo images at a certain video rate, and its output is a mask encoding the static and dynamic elements of the scene, which can be directly used by SLAM or odometry systems.

The first block ({\em Instance Segmentation}) corresponds to the CNN that segments out pixel-wise all the potentially dynamic objects. In our experiments, done using autonomous driving datasets, only cars were segmented as potentially moving. As it will be detailed later, since DOT tracks the mask from frame to frame, this operation does not need to be done at every frame.

The {\em Image processing} block extracts and separates the points belonging to static regions of the image and the points that are in dynamic objects. The camera pose is tracked using only the static part of the scene. From this block, and taking into account the camera pose, the motion of each of the segmented objects is estimated independently ({\em Object tracking}).

The next block ({\em Is the object in motion?}) determines, using geometric criteria, whether the objects labeled as potentially dynamic by the network are indeed moving. This information is used to update the masks encoding the static and dynamic regions of each frame and to feed the linked odometry/SLAM visual system. 

Finally, DOT generates new masks from the estimations of the objects movement ({\em Mask Propagation}), so not every frame needs to be segmented by the network (see Figure \ref{fig:secuencia_algoritmo}). Given the significant computational load of instance segmentation, this can be an relevant advantage of DOT compared to other state-of-the-art methods.

\subsection{Instance Segmentation}

We use the deep network Detectron2 \cite{wu2019detectron2} for the segmentation of all potentially movable instances that are present in an image. The output of the network has been modified to obtain in a single image all the segmentation masks. The image areas that are not classified into the potentially moving categories are given a `background' label and are considered static in the subsequent blocks.

We use the COCO Instance Segmentation baseline model with Mask R-CNN R50-FPN 3x \cite{COCO}\cite{MaskRCNN_github}. The classes have been restricted to those considered as potentially movable, excluding humans since people tracking is beyond the scope of this paper. In case other categories were needed, the network could be fine-tuned using these weights as a starting point or trained from scratch with its own dataset.

In order to consistently track the objects across multiple frames we have included a matching step between the masks computed by DOT and the ones provided by the net. New detections which cannot be paired with to any existing object are used to initialize new instances.

\subsection{Camera and Object Tracking}
From the instance segmentation of the previous step, we aim to estimate the motion of the camera and the dynamic objects. Since the motion of the camera and the motion of the objects are coupled in the images, we make the estimation in a two-step process. First we find the pose of the camera as a relative transformation  ${\color{blue}\mathbf{T}_c} \in \mathbf{SE}(3)$ and then we subtract it to estimate the object motion ${\color{red}\mathbf{T}_o} \in \mathbf{SE}(3)$. 

Our optimization is related to the recent approaches of direct visual odometry and SLAM \cite{engel2017direct}, which aim to find the motion that minimizes a photometric reprojection error.

\textbf{Optimization.} Both for the calculation of camera pose and for the subsequent estimation of object motion, we do Gauss-Newton optimization 

\begin{equation} \label{eq:BA}
        (\textbf{J}^T \boldsymbol{\Sigma}_{r}^{-1}\textbf{J})\textbf{x} = -\textbf{J}^T \boldsymbol{\Sigma}_{r}^{-1}\textbf{r},
\end{equation}

\noindent where $\textbf{J} \in \mathbb{R}^{n\times 6} $ contains the derivatives of the residual function (equations \eqref{eq:efot_cam} and \eqref{eq:efot_obj}) and $\boldsymbol{\Sigma}_{r} \in \mathbb{R}^{n\times n}$ is a diagonal matrix containing the covariances of the photometric residuals 
$\textbf{r} \in \mathbb{R}^{n} $.  The Lie-algebras pose-increments $\widehat{\textbf{x}}_{\mathfrak{se}(3)} \in \mathfrak{se}(3)$, with $\widehat{\cdot}_{\mathfrak{se}(3)}$ being the mapping operator from the vector to the matrix representation of the tangent space \cite{strasdat2012local}, are expressed as a vector $\textbf{x} \in \mathbb{R}^6$. We update the transformations using left matrix multiplication and the exponential map operator $\exp(\cdot)$. Both optimizations are initialized with a constant velocity model and a multi-scale pyramid image to aid convergence.

\textbf{Camera tracking.} The camera motion is estimated using the static scene points $\mathcal{P}$ and multi-view constraints \cite{multiview}, assuming that the camera calibration and points depths are known. The projection of a static point $\mathbf{p} \in \mathcal{P} $ from its pixel coordinates $\mathbf{p}^j$ in the reference frame $F_j$ to its corresponding coordinates $\mathbf{p}^i$ in the frame $F_i$ is as follows:

\begin{equation}
\label{eq:projection2}
\mathbf{p}^i =  \Pi({\color{blue}\mathbf{T}_c}\Pi^{-1}(\mathbf{p}^j,z_j)) ,
\end{equation}

\noindent where 
$\Pi$ and $\Pi^{-1}$ correspond to perspective projection and back-projection models, respectively, and $z_j$ is the depth of the point in the reference frame $F_j$.

The camera pose is optimized by minimizing the photometric reprojection error

\begin{equation}
\label{eq:efot_cam}
\sum_{p \in \mathcal{P}}\Big|\Big|I_j(\mathbf{p}^j)- I_i( \Pi(\exp({\color{blue}\widehat{\textbf{x}}_{\mathfrak{se}(3)}}){\color{blue}\mathbf{T}_c}\Pi^{-1}(\mathbf{p}^j,z_j)))\Big|\Big|_\gamma ,
\end{equation}

\noindent which is computed as the sum of all intensity differences between points in their reference frame and their projection into the frame being tracked. We use the Huber norm $\gamma$.

\textbf{Object tracking.} Once  $\color{blue}\mathbf{T_{c}}$ has been estimated, the pose of each potentially dynamic object can be estimated analogously by using the image points $\mathcal{Q}$ belonging to such object. Modelling the potentially dynamic object as a solid with pose $\color{red}\mathbf{T_{o}}$, the projection of each point $\widetilde{\mathbf{p}}$ in the frame $F_j$ to its coordinates in frame $F_i$ is: 

\begin{equation}
\label{eq:projection3}
\widetilde{\mathbf{p}}^i =  \Pi({\color{blue}\mathbf{T}_c}{\color{red}\mathbf{T}_o}\Pi^{-1}(\widetilde{\mathbf{p}}^j,z_j)) .
\end{equation}

Analogously to equation \ref{eq:efot_cam}, we estimate  $\color{red}\mathbf{T}_o$ by minimizing the following photometric reprojection error 

\begin{equation}
\label{eq:efot_obj}
\sum_{\widetilde{p} \in \mathcal{Q}} \Big|\Big|I_j(\widetilde{\mathbf{p}}^j)- I_i(\Pi({\color{blue}\mathbf{T}_c}\exp({\color{red}\widehat{\textbf{x}}_{\mathfrak{se}(3)}}){\color{red}\mathbf{T}_o}\Pi^{-1}(\widetilde{\mathbf{p}}^j,z_j)))\Big|\Big|_\gamma .
\end{equation}

\label{sec:motionState}

\begin{figure*}[t]
\begin{center}
\includegraphics[width=180mm]{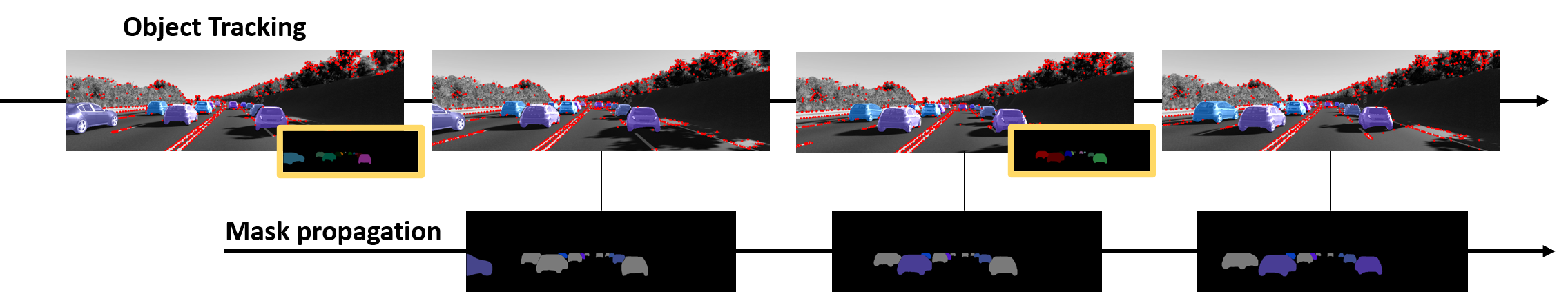}
\end{center}
  \caption{\textbf{Sample of a segment of the computation flow}. The upper row shows DOT estimating the tracking of the camera and the objects. Note how the segmentation masks from the network (yellow frames) are not necessary in all frames. The lower row shows the segmantic masks generated by DOT that encode the motion classification: in motion (color), static (black) and not observed (gray).}
    \label{fig:secuencia_algoritmo}
\end{figure*}

\subsection{Tracking quality, outliers and occlusions}

\begin{algorithm}[t]
\caption{Dynamic Object Tracking
\label{alg:objectTracking}}
\begin{algorithmic}[1]
    \Function{Object Tracking}{$\mathcal{P}, \mathcal{Q}, \mathcal{O}$}
    \State \Comment{$\mathcal{P}$ = static points} 
    \State \Comment{$\mathcal{Q}$ = dynamic points}
    \State \Comment{$\mathcal{O}$ = set of objects}
    \Let {mask} {$\emptyset$} \Comment{Dynamic mask to be computed}
    \State 
    \Let {\{$T_c$ , $\phi_c$\}}{track camera ($\mathcal{P}$)} \Comment{Camera Tracking}
    \If{$\phi_c < th_{\phi} $} \Return{$\emptyset$} \EndIf
    \State
    \For{$\textrm{object} \textrm{ \textbf{in} $\mathcal{O}$}$} \Comment{Object Tracking}
        \If{is visible (object , $T_c$)}
            \Let{\{$T_o$ , $\phi_o$\}}{track object ($T_c$ , $\mathcal{Q}_o$ , mask)}
            \If{$\phi_o < th_{\phi} $} 
            \textbf{break}
            \EndIf
            \Let{object}{outlier rejection ($\phi_o$)}
            \Let{mask}{update mask (object)}
            \Let{mask}{is object moving? (object)}
        \EndIf
        
    \EndFor
    \State 
    \State \Return{mask}

    \EndFunction
\end{algorithmic}
\end{algorithm}

Occlusions, changes in lighting conditions and segmentation errors have a significant effect in the accuracy of the objects and camera poses. As seen in algorithm \ref{alg:objectTracking}, we developed several strategies that we apply after the object tracking step to reduce their impact. 

\textbf{Tracking quality.} The appearance of dynamic objects changes significantly, producing high tracking errors. We used the Pearson's correlation coefficient $\phi_o \in [-1, 1]$ to model appearance similarity. This metric reflects the degree of linear correlation between the reference intensities of the points and their corresponding estimates, hence being invariant to changes in gain and offset. Note that this metric can also be applied to camera tracking $\phi_c$, although changes in the appearance of the background are usually less pronounced. 

\begin{figure}[]
  \centering
\includegraphics[width=85mm]{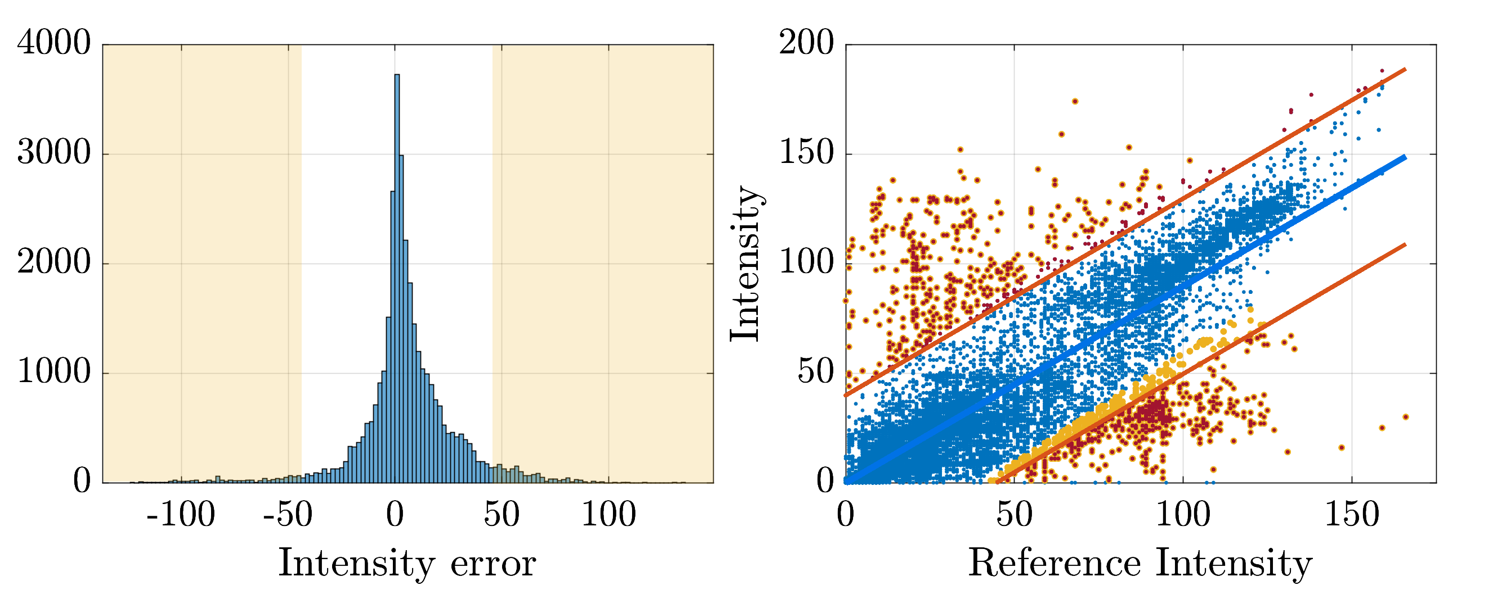}
  \caption{\textbf{Outlier rejection}. \textbf{Left:} histogram of photometric errors for an object. The shaded area corresponds to the points removed with a constant threshold. \textbf{Right:} Linear relation between intensities. Note the different points labeled as outliers by absolute (yellow) and relative (red) thresholds due to the changing photometry.}
    \label{fig:pearson}
\end{figure}

\textbf{Outlier rejection.} A common approach to detect outliers is defining an absolute threshold to the photometric error \eqref{eq:efot_cam} \eqref{eq:efot_obj}. More sophisticated works \cite{engel2017direct} adapt it according to the median residual, the motion blur or the lighting changes. As shown in Figure \ref{fig:pearson}, we propose to set the threshold relative to the linear relation between intensities, so the errors are independent to photometric changes in the image.  

\textbf{Occlusions.} The dynamic objects might occlude each other. Removing the occluded parts as outliers was not sufficient in our experiments. We implemented a strategy consisting of tracking the objects from the closest to the farthest, updating their respective masks sequentially. In this manner, we update in every iteration the points of the more distant objects that have been occluded by closer ones.

\subsection{Is the object in motion?}

This block receives as input the transformation matrices of the camera, $\mathbf{T}_c$ and the objects, $\mathbf{T}_o$, and estimates whether the objects are moving or not. Its output, to be used by SLAM or odometry systems, are the masks that store the areas of the image occupied by dynamic objects and whether they are in motion or not. The masks are obtained by projecting the pixels of each object into the new frame using $\mathbf{T}_c$ and $\mathbf{T}_o$ estimated in the previous step.

\begin{figure}[]
  \centering
\includegraphics[width=80mm]{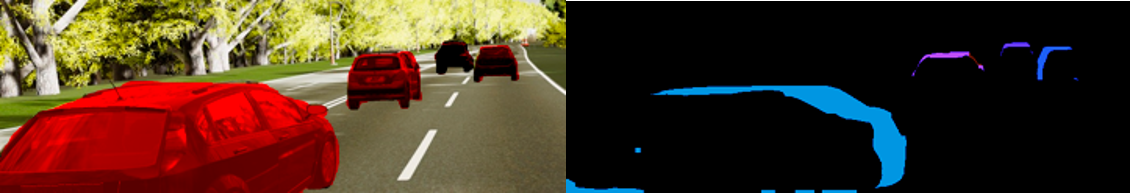}
  \caption{\textbf{Disparity vs Entropy}. Comparison of the dynamic disparities produced by different objects in motion. Note how observations with high entropy values (brighter red) produce larger shifts of image pixels.}
    \label{fig:disparityEntropy}
\end{figure}

Observing the object motion directly in $\mathbf{T}_o$ generates, due to the propagated image noise, difficulties in establishing absolute thresholds that determine whether an object is in motion. In this work we chose to observe the motion of the objects using 2D image measurements. We denote our metric as \textit{dynamic disparity}, being the distance in pixels between the projection of the point as if it were static $\mathbf{p}^i$ and its actual projection $\widetilde{\mathbf{p}}^i$. For each object we compute the median of the dynamic disparities of its points $\widetilde{\mathbf{p}} \in \mathcal{Q}$:

\begin{equation}
\label{eq:disparity_dinamic}
d_d = \textrm{med} \big\{\left |\left |\mathbf{p}^i,\widetilde{\mathbf{p}}^i\right |\right | , \forall \widetilde{\mathbf{p}}\in \mathcal{Q} \big\}.
\end{equation}

The 3D motion of a point produces different image motions depending on 1) its image coordinates, 2) its depth, and 3) the relative angle between the directions of the object and the camera motions.  

From the non-linear pose optimization (see eq. \eqref{eq:BA}) we can derive the uncertainty in the estimation of the motion of the object  $\boldsymbol{\Sigma}_{x} =  (\textbf{J}^T \boldsymbol{\Sigma}_{{ r}}^{-1}\textbf{J})^{-1}$. Assuming a k-dimensional Gaussian distribution, its differential entropy is:

\begin{equation} \label{eq:entropy}
    H({\textbf{x}_o}) = \frac{1}{2}\log((2\pi e)^{k}|\boldsymbol{\Sigma}_{ x_o}|).
\end{equation}

The differential entropy can be seen as the pose uncertainty derived from the photometric residuals minimization. In other words, observations of three-dimensional motions with high entropy values will result in larger shifts of image pixels (see Figure \ref{fig:disparityEntropy}). On the other hand, observations with low entropy will produce small image disparities.

Based on this, the algorithm for classifying the movement of objects works as follows. We compare dynamic disparities \eqref{eq:disparity_dinamic} against a variable threshold $\Delta d = f(H(x))$ that grows smoothly with the entropy. We label as ``in motion'' all those objects whose dynamic disparity exceeds this threshold ($d_d  > \Delta d $). For every value below an entropy threshold $H_{min}$ we assume the object motion cannot be observed. Therefore, labeling an object as static requires that the motion is observable ($H(x)  > H_{min}$) and that the median of the dynamic disparity is less than the variable threshold  ($d_d < \Delta d$).

While selecting the optimal functional formulation would require further study, this expression meets the requirements and has shown good results in this work (see section \ref{sec:evaluation}). Figure \ref{fig:secuencia_algoritmo} is an example of the mask propagated by DOT.  Objects labeled as ``in motion'' are represented in colour, while those labeled as ``static'' disappear in black. The cars represented in gray are those which cannot be determined as being static neither dynamic.

\subsection{Mask propagation}
\label{sec::maskPropagation}
DOT exploits the two segmentation masks available in each frame: one produced by the neural network and other propagated from the previous frame. Warping one segmentation into the other allows to robustly relate instances found in different frames into the same 3D object.

\textbf{State propagation.} Relating new semantic instances to pre-existing objects allows us to predict their motion (which is critical for fast moving objects). In addition, it is possible to keep the classification of the motion in the case of an object moving to a position where the motion is not observable (see Section \ref{sec:motionState}). 

\textbf{Independent segmentation.} Our proposal allows the propagation of semantic segmentation masks from an initial seed over time and space, eliminating the need for segmenting every frame. Running the neural network at a lower frequency makes real-time object tracking easier in low-end platforms. As further benefit, DOT is able to fill in the gaps in which the network temporarily loses the instantiation of an object between consecutive images.

\section{Experimental results}
Although the potential applications of DOT cover a wide spectrum ranging from object detection to augmented reality or autonomous driving, in this paper we provide an intensive evaluation to demonstrate to what extent ``knowing the movement of objects'' can improve the accuracy of a SLAM system.

\subsection{Evaluation against baselines}
\label{sec:evaluation}
\textbf{Baselines.} Our experiments estimate the camera trajectory using a state-of-the-art SLAM system in three different configurations. Specifically, we use ORB-SLAM2 \cite{Mur_Artal_2017}, with its RGB-D and stereo implementations.  The three configurations designed to evaluate DOT are:

\textit{No masks}: ORB-SLAM2 is run using the authors' implementation on unmodified images. A rigid scene is assumed, so all the points in the images (including those belonging to moving objects) can be selected by ORB-SLAM2.

\textit{DOT masks}: ORB-SLAM2 receives as input, in addition to the images, the dynamic object masks containing potentially dynamic objects currently in motion. We modified ORB-SLAM2 so that it does not extract points from such moving objects.     

\textit{All masks}: ORB-SLAM2 receives all the masks obtained by the instance segmentation network. In this configuration, all potentially dynamic objects are removed without checking if they are actually moving or not.

\begin{figure*}[]
  \centering
\includegraphics[width=170mm]{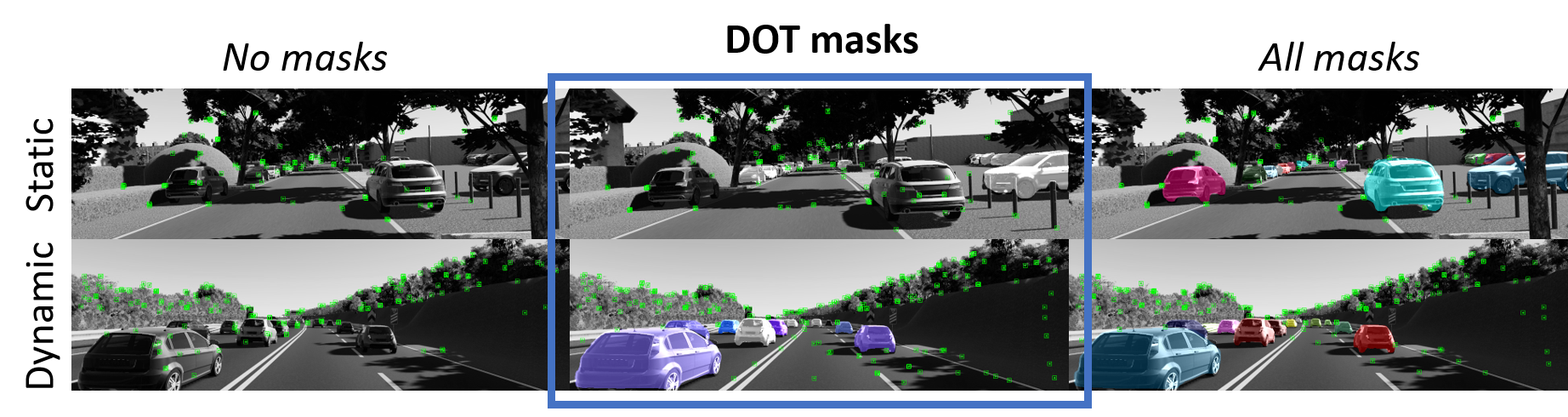}  
  \caption{\textbf{Scene content adaptation}. Sample results for the three studied configurations. \textbf{Left}: \textit{No masks}. \textbf{Centre}: \textit{DOT masks}.  \textbf{Right}: \textit{All masks}.
  The top row shows a static scene in which the \textit{All masks} setting discards all points of the static objects that can aid to tracking accuracy. In contrast, the bottom row shows how the \textit{No masks} configuration allows to extract points on moving objects that may cause the system to fail. Both are cases in which the lack of understanding of the scene deteriorate the performance of SLAM. DOT successfully identifies the parked cars as static and the moving ones as dynamic. Note how DOT achieves a trade-off between those two opposing scenarios by estimating the actual motion state of the objects that results in a better estimation of the trajectory.}
    \label{fig:comparacion_metodos}
\end{figure*}

\begin{table}[h]
\huge
\begin{center}
\centering
\resizebox{0.48\textwidth}{!}{%
\begin{tabular}{c|c c c p{0.01cm} c c c}
\toprule
                       & \multicolumn{3}{c}{ATE {[}m{]}}                                                        & \multicolumn{1}{p{0.01cm}}{}& \multicolumn{3}{c}{$\text{ATE/}\text{ATE}_{\text{best}}$}                                                                                                     \\
\multirow{-2}{*}{Seq.} & \multicolumn{1}{c}{
\begin{tabular}[c]{@{}c@{}}No\\ {masks}\end{tabular}}

                        &  \multicolumn{1}{c}{DOT}                       & \multicolumn{1}{c}{
\begin{tabular}[c]{@{}c@{}}All\\ {masks}\end{tabular}} &\multicolumn{1}{p{0.01cm}}{}& \multicolumn{1}{c}{
\begin{tabular}[p{3cm}]{@{}c@{}}No\\ {masks}\end{tabular}}                                         & \multicolumn{1}{c}{DOT}                                               & \multicolumn{1}{c}{
\begin{tabular}[p{3cm}]{@{}c@{}}All\\ {masks}\end{tabular}}                  \\ \hline
01                     & \multicolumn{1}{c}{\textbf{1.10}} & \multicolumn{1}{c}{1.14}          & \multicolumn{1}{c}{1.38}          &\multicolumn{1}{p{0.1cm}}{}& \multicolumn{1}{c}{\cellcolor[HTML]{A0CE62}1.00} & \multicolumn{1}{c}{\cellcolor[HTML]{B5D469}1.04} & \multicolumn{1}{c}{\cellcolor[HTML]{F6C770}1.26} \\
02                     & \multicolumn{1}{c}{0.16}          & \multicolumn{1}{c}{0.14}          & \multicolumn{1}{c}{\textbf{0.10}} &\multicolumn{1}{p{0.1cm}}{}& \multicolumn{1}{c}{\cellcolor[HTML]{CA2012}1.60} & \multicolumn{1}{c}{\cellcolor[HTML]{E07542}1.43} & \multicolumn{1}{c}{\cellcolor[HTML]{A0CE62}1.00} \\
06                     & \multicolumn{1}{c}{0.11}          & \multicolumn{1}{c}{\textbf{0.07}} & \multicolumn{1}{c}{0.08}          &\multicolumn{1}{p{0.1cm}}{}& \multicolumn{1}{c}{\cellcolor[HTML]{C10000}1.67} & \multicolumn{1}{c}{\cellcolor[HTML]{A0CE62}1.00} & \multicolumn{1}{c}{\cellcolor[HTML]{FFEB84}1.18} \\
18                     & \multicolumn{1}{c}{4.77}          & \multicolumn{1}{c}{\textbf{1.00}} & \multicolumn{1}{c}{1.50}          &\multicolumn{1}{p{0.1cm}}{}& \multicolumn{1}{c}{\cellcolor[HTML]{C10000}4.79} & \multicolumn{1}{c}{\cellcolor[HTML]{A0CE62}1.00} & \multicolumn{1}{c}{\cellcolor[HTML]{D6502D}1.51} \\
20                     & \multicolumn{1}{c}{29.42}         & \multicolumn{1}{c}{\textbf{9.12}} & \multicolumn{1}{c}{13.54}         &\multicolumn{1}{p{0.1cm}}{}& \multicolumn{1}{c}{\cellcolor[HTML]{C10000}3.23} & \multicolumn{1}{c}{\cellcolor[HTML]{A0CE62}1.00} & \multicolumn{1}{c}{\cellcolor[HTML]{D95932}1.49} \\ \hline
\multicolumn{1}{c}{$\overline{\varepsilon}_{norm}$}                 & \multicolumn{1}{c}{192.6\%}         & \multicolumn{1}{c}{100.0\%} & \multicolumn{1}{c}{137.8\%} &\multicolumn{1}{p{0.01cm}}{}& \multicolumn{1}{c}{} & \multicolumn{1}{p{2.5cm}}{} & \multicolumn{1}{c}{} \\ \bottomrule

\end{tabular}%
}
\end{center}
\caption{
DOT against baselines (\textit{No masks} and  \textit{All masks}) in V-KITTI. \textbf{Left:} ATE {[}m{]}. \textbf{Right:} ATE over best ATE per sequence.}
\label{tab:vkitti_results}

\end{table}

\begin{table}[h]
\centering
\huge
\begin{center}
\resizebox{0.48\textwidth}{!}{%
\begin{tabular}{p{1cm}|p{1cm}p{1.5cm}p{2cm}p{0.05cm}p{2cm}p{2cm}p{2cm}}
\toprule
                       & \multicolumn{3}{c}{ATE {[}m{]}}                        &\multicolumn{1}{c}{}& \multicolumn{3}{c}{$\text{ATE/}\text{ATE}_{\text{best}}$}\\
\multirow{-2}{*}{Seq.} & \multicolumn{1}{c}{
\begin{tabular}[c]{@{}c@{}}No\\ {masks}\end{tabular}}

                        &  \multicolumn{1}{c}{DOT}                       & \multicolumn{1}{c}{
\begin{tabular}[c]{@{}c@{}}All\\ {masks}\end{tabular}} &\multicolumn{1}{p{0.01cm}}{}& \multicolumn{1}{c}{
\begin{tabular}[p{3cm}]{@{}c@{}}No\\ {masks}\end{tabular}}                                         & \multicolumn{1}{c}{DOT}                                               & \multicolumn{1}{c}{
\begin{tabular}[p{3cm}]{@{}c@{}}All\\ {masks}\end{tabular}}                  \\ \hline

\multicolumn{1}{c|}{0}                      & \multicolumn{1}{c}{\textbf{1.77}} & \multicolumn{1}{c}{1.80}          & \multicolumn{1}{c}{2.08}          & & \multicolumn{1}{c}{\cellcolor[HTML]{A0CE62}1.00} & \multicolumn{1}{c}{\cellcolor[HTML]{B4D469}1.02} & \multicolumn{1}{c}{\cellcolor[HTML]{E79051}1.18} \\

\multicolumn{1}{c|}{1}                      & \multicolumn{1}{c}{\textbf{6.37}} & \multicolumn{1}{c}{7.71}          & \multicolumn{1}{c}{8.45}          & & \multicolumn{1}{c}{\cellcolor[HTML]{A0CE62}1.00} & \multicolumn{1}{c}{\cellcolor[HTML]{DE6E3E}1.21} & \multicolumn{1}{c}{\cellcolor[HTML]{C00000}1.33} \\
\multicolumn{1}{c|}{2}                      & \multicolumn{1}{c}{3.72}          & \multicolumn{1}{c}{\textbf{3.70}} & \multicolumn{1}{c}{3.84}          & &\multicolumn{1}{c}{\cellcolor[HTML]{A6CF64}1.01} & \multicolumn{1}{c}{\cellcolor[HTML]{A0CE62}1.00} & \multicolumn{1}{c}{\cellcolor[HTML]{CBDB71}1.04} \\
\multicolumn{1}{c|}{3}                      & \multicolumn{1}{c}{0.40}          & \multicolumn{1}{c}{0.40}          & \multicolumn{1}{c}{0.40}          & &\multicolumn{1}{c}{\cellcolor[HTML]{A0CE62}1.00} & \multicolumn{1}{c}{\cellcolor[HTML]{AED267}1.01} & \multicolumn{1}{c}{\cellcolor[HTML]{A0CE62}1.00} \\
\multicolumn{1}{c|}{4}                      & \multicolumn{1}{c}{0.27}          & \multicolumn{1}{c}{0.26}          & \multicolumn{1}{c}{\textbf{0.24}}& &\multicolumn{1}{c}{\cellcolor[HTML]{F4C26D}1.12} & \multicolumn{1}{c}{\cellcolor[HTML]{FDE380}1.09} & \multicolumn{1}{c}{\cellcolor[HTML]{A0CE62}1.00} \\
\multicolumn{1}{c|}{5}                      & \multicolumn{1}{c}{0.40}          & \multicolumn{1}{c}{\textbf{0.39}} & \multicolumn{1}{c}{0.45}          & &\multicolumn{1}{c}{{\cellcolor[HTML]{C1D86E}1.03}} & \multicolumn{1}{c}{\cellcolor[HTML]{A0CE62}1.00} & \multicolumn{1}{c}{\cellcolor[HTML]{EFB063}1.14} \\
\multicolumn{1}{c|}{6}                     & \multicolumn{1}{c}{\textbf{0.63}} &\multicolumn{1}{c}{0.68}          & \multicolumn{1}{c}{0.67}          & &\multicolumn{1}{c}{\cellcolor[HTML]{A0CE62}1.00} & \multicolumn{1}{c}{\cellcolor[HTML]{FFE983}1.08} & \multicolumn{1}{c}{\cellcolor[HTML]{EDE57D}1.07} \\
\multicolumn{1}{c|}{7}                      & \multicolumn{1}{c}{0.52}          & \multicolumn{1}{c}{\textbf{0.51}} & \multicolumn{1}{c}{\textbf{0.51}} & &\multicolumn{1}{c}{\cellcolor[HTML]{A7D064}1.01} & \multicolumn{1}{c}{\cellcolor[HTML]{A0CE62}1.00} & \multicolumn{1}{c}{\cellcolor[HTML]{A0CE62}1.00 }\\
\multicolumn{1}{c|}{8}                      & \multicolumn{1}{c}{\textbf{3.04}} & \multicolumn{1}{c}{3.24}          & \multicolumn{1}{c}{3.78}          & &\multicolumn{1}{c}{\cellcolor[HTML]{A0CE62}1.00} & \multicolumn{1}{c}{\cellcolor[HTML]{EEE57E}1.07} & \multicolumn{1}{c}{\cellcolor[HTML]{D6502D}1.24} \\
\multicolumn{1}{c|}{9}                      & \multicolumn{1}{c}{2.65}          & \multicolumn{1}{c}{\textbf{0.98}} & \multicolumn{1}{c}{3.80}          & & \multicolumn{1}{c}{\cellcolor[HTML]{C10000}2.71} & \multicolumn{1}{c}{\cellcolor[HTML]{A0CE62}1.00} & \multicolumn{1}{c}{\cellcolor[HTML]{B12318}3.89} \\
\multicolumn{1}{c|}{10}                     & \multicolumn{1}{c}{\textbf{1.23}} & \multicolumn{1}{c}{1.29}          & \multicolumn{1}{c}{1.26}          & & \multicolumn{1}{c}{\cellcolor[HTML]{A0CE62}1.00} & \multicolumn{1}{c}{\cellcolor[HTML]{D5DE75}1.05} & \multicolumn{1}{c}{\cellcolor[HTML]{BCD66C}1.02} \\ \hline

\multicolumn{1}{c}{$\overline{\varepsilon}_{norm}$}                 & \multicolumn{1}{c}{112.7\% }         & \multicolumn{1}{p{2.0cm}}{100.0\%} & \multicolumn{1}{c}{130.3\%} &\multicolumn{1}{p{0.1cm}}{}& \multicolumn{1}{p{2.0cm}}{} & \multicolumn{1}{p{2.0cm}}{} & \multicolumn{1}{p{2.0cm}}{} \\ \bottomrule
\end{tabular}%
}
\end{center}
\caption{DOT against baselines (\textit{No masks} and \textit{All masks}) in KITTI \textit{Odometry}. \textbf{Left:} ATE {[}m{]}. \textbf{Right:} ATE over best ATE per sequence.}
\label{tab:kitti_ate_odometry_results}
\end{table}

\begin{table}[h]
\Huge
\begin{center}
\centering
\resizebox{0.49\textwidth}{!}{%
\begin{tabular}{c|c c c  p{0.001cm}p{1cm}p{1cm}p{1cm}}
\toprule
                       & \multicolumn{3}{c}{ATE {[}m{]}}                        &\multicolumn{1}{c}{}& \multicolumn{3}{c}{$\text{ATE/}\text{ATE}_{\text{best}}$}\\
\multirow{-2}{*}{Seq.} & \multicolumn{1}{c}{
\begin{tabular}[c]{@{}c@{}}No\\ {masks}\end{tabular}}

                        &  \multicolumn{1}{c}{DOT}                       & \multicolumn{1}{c}{
\begin{tabular}[c]{@{}c@{}}All\\ {masks}\end{tabular}} &\multicolumn{1}{p{0.01cm}}{}& \multicolumn{1}{c}{
\begin{tabular}[p{3cm}]{@{}c@{}}No\\ {masks}\end{tabular}}                                         & \multicolumn{1}{c}{DOT}                                               & \multicolumn{1}{c}{
\begin{tabular}[p{3cm}]{@{}c@{}}All\\ {masks}\end{tabular}}                  \\ \hline
0926-0009              & \multicolumn{1}{c}{\textbf{1.23}} & \multicolumn{1}{c}{1.24}          & \multicolumn{1}{c}{1.44}          && \multicolumn{1}{c}{\cellcolor[HTML]{92D050}1.00}  & \multicolumn{1}{c}{\cellcolor[HTML]{A9D55B}1.01} & \multicolumn{1}{c}{\cellcolor[HTML]{EB9F5A}1.17} \\
0926-0013              & \multicolumn{1}{c}{\textbf{0.26}} & \multicolumn{1}{c}{\textbf{0.26}} & \multicolumn{1}{c}{0.27}          && \multicolumn{1}{c}{\cellcolor[HTML]{92D050}1.00}  & \multicolumn{1}{c}{\cellcolor[HTML]{9BD254}1.00} & \multicolumn{1}{c}{\cellcolor[HTML]{D8E171}1.03} \\
0926-0014              & \multicolumn{1}{c}{0.86}          & \multicolumn{1}{c}{0.82}          & \multicolumn{1}{c}{\textbf{0.78}} && \multicolumn{1}{c}{\cellcolor[HTML]{F5C46F}1.11}  & \multicolumn{1}{c}{\cellcolor[HTML]{FDE480}1.06} & \multicolumn{1}{c}{\cellcolor[HTML]{92D050}1.00} \\
0926-0051              & \multicolumn{1}{c}{0.37}          & \multicolumn{1}{c}{\textbf{0.36}} & \multicolumn{1}{c}{0.37}          && \multicolumn{1}{c}{\cellcolor[HTML]{C7DD69}1.02}  & \multicolumn{1}{c}{\cellcolor[HTML]{92D050}1.00} & \multicolumn{1}{c}{\cellcolor[HTML]{C7DD69}1.02} \\
0926-0101              & \multicolumn{1}{c}{\textbf{8.66}} & \multicolumn{1}{c}{10.26}         & \multicolumn{1}{c}{12.37}         && \multicolumn{1}{c}{\cellcolor[HTML]{92D050}1.00}  & \multicolumn{1}{c}{\cellcolor[HTML]{E99856}1.18} & \multicolumn{1}{c}{\cellcolor[HTML]{C00000}1.43} \\
0929-0004              & \multicolumn{1}{c}{0.32}          & \multicolumn{1}{c}{\textbf{0.30}} & \multicolumn{1}{c}{\textbf{0.30}} && \multicolumn{1}{c}{\cellcolor[HTML]{FAD97A}1.08}  & \multicolumn{1}{c}{\cellcolor[HTML]{D5E070}1.03} & \multicolumn{1}{c}{\cellcolor[HTML]{92D050}1.00} \\
1003-0047              & \multicolumn{1}{c}{13.81}         & \multicolumn{1}{c}{\textbf{1.25}} & \multicolumn{1}{c}{2.23}          && \multicolumn{1}{c|}{\cellcolor[HTML]{C00000}11.01} & \multicolumn{1}{c}{\cellcolor[HTML]{92D050}1.00} & \multicolumn{1}{c}{\cellcolor[HTML]{C00000}1.78} \\ \hline
\multicolumn{1}{c}{$\overline{\varepsilon}_{norm}$}              & \multicolumn{1}{c}{242.3\%}         & \multicolumn{1}{c}{100.0\%} & \multicolumn{1}{c}{115.9 \%}          &\multicolumn{1}{p{0.001cm}}{}& \multicolumn{1}{c}{} & \multicolumn{1}{p{3cm}}{} & \multicolumn{1}{c}{} \\ \bottomrule
\end{tabular}%
}
\end{center}
\caption{DOT against baselines (\textit{No masks} and \textit{All masks}) in KITTI \textit{Raw}. \textbf{Left:} ATE {[}m{]}. \textbf{Right:} ATE over best ATE per sequence.}
\label{tab:kitti_tracking_results}
\end{table}

\textbf{Sequence subsets.} We evaluate the above configurations in three sequence subsets from the KITTI Vision Benchmark Suite \cite{KITTI}, containing stereo sequences of urban and road scenes recorded from a car and used for research in autonomous driving. We use Virtual KITTI \cite{VKITTI} \cite{VKITTI2}, a synthetic dataset composed of 5 sequences virtually cloned from KITTI \cite{KITTI}, KITTI \textit{Odometry}, a predefined subset of sequences specially designed for the development and evaluation of visual odometry systems, and a selection of sequences chosen from the \textit{raw} section of KITTI because of their high number of moving objects \cite{huang2020clustervo}. 

We run the RGB-D version of ORB-SLAM2 in Virtual KITTI, as synthetic depth images are provided, while for the other subsets we run the stereo version of ORB-SLAM2 over the color stereo pairs. The ground truth for the real sequences is given by an accurate GPS localization system. 

\textbf{Evaluation metrics.} As it is standard when evaluating real-time SLAM, in order to take into account non-deterministic effects, we run each configuration 10 times per sequence and report median values. All the experiments were run in a laptop with an Intel Core i5 processor and 8GB of RAM memory.
 
We report the absolute trajectory error (ATE) as proposed in \cite{sturm2012benchmark}, which is the root-mean square error (RMSE) of the estimated position of all frames with respect to the GPS ground truth after both trajectories have been aligned. For an easier comparison between DOT and the other two configurations, we report the average of the errors normalized by the value we obtained with DOT on each sequence $\overline{\varepsilon}_{norm}=\frac{1}{n}\sum_{i=0}^{n}\frac{\varepsilon_i }{\varepsilon_{DOT}}$.


The right columns in Tables \ref{tab:vkitti_results}, \ref{tab:kitti_ate_odometry_results}, \ref{tab:kitti_tracking_results} show the ATE normalized by the best ATE in each sequence among the three configurations. Thus, a value equal to 1 identifies the best result, while values $>1$ are indicative of poorer performance. The color scale indicates the trade off of the errors between the best result (green) and the worst (red).

\textbf{Tracking accuracy.} The ATE in Table \ref{tab:vkitti_results}, corresponding to the V-KITTI sequences, show an accuracy improvement of 92.6\% and 37.8\% of our system with respect to the \textit{No masks} and \textit{All masks} configurations, respectively. In addition, DOT scores best for 3 of the 5 sequences evaluated.

Table \ref{tab:kitti_ate_odometry_results} contains the ATE results for 11 trajectories of KITTI \textit{Odometry} evaluated with the three different configurations. DOT obtains in this case an overall performance which is 12.7\% and 30.3\% better than \textit{No masks} and \textit{All masks}, respectively. Compared to V-KITTI, this group of sequences contains less dynamic elements, so the use of masks is even detrimental. According to the dataset specifications, the ground truth camera poses collected by the GPS are accurate to within 10 cm. Therefore, no significant differences exist between the three configurations in sequences 3, 4, 5, 6, 7 and 10. This is thought to be a consequence of the small number of moving objects, as well as of the rich texture of the images, which provides a large number of static points for estimating the camera motion.

The differences between sequences and methods are more evident with the last set of sequences shown in Table \ref{tab:kitti_tracking_results}, characterized by an abundance of moving objects. Overall, DOT achieves improvements of 142.3\% in ATE accuracy over \textit{No masks} and 15.9 \% over the \textit{All masks} method. Again, note how discarding dynamic objects in sequence 1003-0047 reduces significantly the tracking errors. The sequences 0926-0009, 0929-0004 and 1003-0047 were cloned to generate the V-KITTI synthetic sequences (1, 18 and 20). As expected, since the scenes contents are identical, so is the qualitative analysis of the results. 

The color scale used in Tables \ref{tab:vkitti_results}, \ref{tab:kitti_ate_odometry_results}, \ref{tab:kitti_tracking_results} shows how DOT tends to approach to the best solution when it is not the most accurate trajectory (green). This proves that, while the use of masks may be convenient, the accuracy is significantly improved if only the objects that have been verified to be in motion are removed. These results demonstrate that DOT achieves consistently a good performance both for static and dynamic scenes.

\textbf{ Scene content adaptation.} Figure \ref{fig:comparacion_metodos} illustrates two scenarios that affect the SLAM accuracy in a scene with dynamic objects. The lower row shows a road where all the vehicles are in motion (Seq. 20 in Table \ref{tab:vkitti_results}). The high dynamism of all the vehicles in the scene violates the rigidity assumption of ORB-SLAM2, and makes the system fail. Similarly, moving objects in sequence 18 (Table \ref{tab:vkitti_results}) causes tracking failure of ORB-SLAM2 in 6 out of 10 trials (only 56\% of the trajectory could be estimated in those cases).

The upper row shows an urban scene with several cars parked on both sides of the road (Seq. 01 in Table \ref{tab:vkitti_results}). Contrary to the previous case, the worst configuration is using all the segmentation masks since a large number of points with high information content are removed for tracking. ATE results in Table \ref{tab:vkitti_results} for this sequence shows that extracting points from a larger area results in a better accuracy of the estimated trajectory.

Summing up, notice how not using dynamic object masks increases the trajectory error due to matching points on moving objects. However, applying masks without verifying if the object is in motion discards a high amount of information, especially when a large part of the scene is occupied by vehicles. DOT achieves a trade-off between those two opposing scenarios by estimating the actual motion state of the objects in order to get higher tracking robustness and accuracy. 

\textbf{Loop closure.} Not all differences in trajectory accuracy are due to poor tracking performance. The loop closure module of ORB-SLAM2 reduces the drift and therefore also the inaccuracies produced by dynamic objects or by the removal of parked vehicles. We have observed that ORB-SLAM2 running with \textit{DOT masks} is able to close the loop 6 out of 10 runs in sequence 9 of KITTI \textit{Odometry} (see Table \ref{tab:kitti_ate_odometry_results}), while none was identified when using \textit{All masks}. This results in a broader error variability.

\begin{figure}[]
  \centering
\includegraphics[width=80mm]{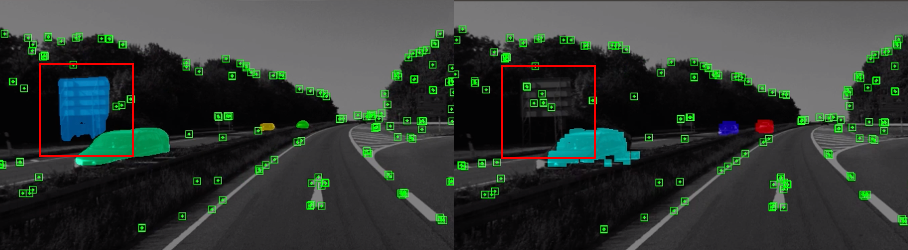}
  \caption{\textbf{Segmentation error}. Comparison between \textit{All masks} and \textit{DOT masks}. Notice that a wrong segment from Detectron2 (the sign in the red square is assigned a car label) is correctly classified as static by DOT.}
    \label{fig:errores_red}
\end{figure}

\textbf{Segmentation errors.} Compared to other approaches, DOT is capable of alleviating segmentation errors. Neural networks sometimes mislabel static objects (\eg, traffic signs or buildings) as dynamic, DOT corrects this error by re-tagging the object as static  (see Figure \ref{fig:errores_red}). As another example, when the network does not fire in one of the sequence frames, DOT is able to fill the gap by propagating the object mask.

\subsection{Mask propagation}

\begin{figure}[]
  \centering
\includegraphics[width=85mm]{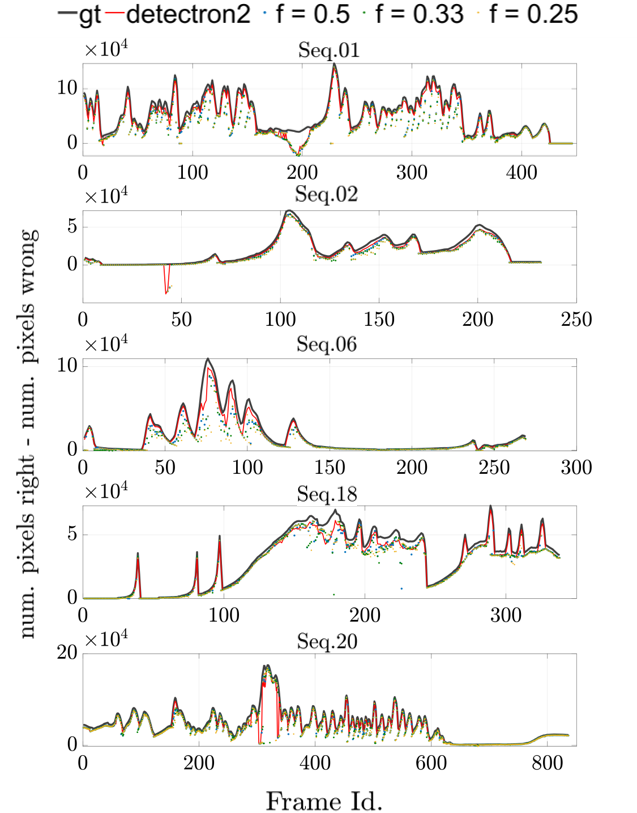}
  \caption{\textbf{Mask propagation}. We show for each frame of V-KITTI the number of correctly labeled pixels minus mislabeled ones respect to the ground truth (black), when DOT uses all masks from Detectron2 (red), 50\% (blue), 33\% (yellow) and 25\% (green).}
    \label{fig:segmentationExp}
\end{figure}

As explained in section \ref{sec::maskPropagation}, our approach allows to reduce the frequency of network segmentation by propagating pre-existing masks in the intermediate frames. Figure \ref{fig:segmentationExp} shows the number of correctly labeled pixels minus mislabeled ones (ground truth in black) on every frame of V-KITTI when DOT uses 100\% of Detectron2 segmentations (red),  50\% (blue), 33\% (yellow) and 25\% (green). Note how the masks stay accurate when being propagated except when tracking failures occur or a moving object enters the scene between segmentations (see also intersection over union on V-KITTI in Table \ref{tab:maskPropagation}). We believe this result may be helpful for real time object tracking specially for high frequency image streams.
\begin{table}[h]
\begin{center}
 \begin{tabular}{|c|| c c c c c|} 
 \hline
     Rate &Seq01 & Seq02 & Seq06 & Seq18  & Seq20  \\ [0.5ex] 
 \hline \hline
 1.0 &  0.88 & 0.88 & 0.84 & 0.90 & 0.89 \\ 
 0.5 &  0.74 & 0.83 & 0.67 & 0.85 & 0.84 \\ 
 0.33 & 0.72 & 0.80 & 0.60 & 0.85 & 0.81 \\ 
 0.25 & 0.69 & 0.78 & 0.55 & 0.84 & 0.81  \\

 \hline
 
\end{tabular}
\end{center}
\caption{\textbf{Intersection over union} in the V-KITTI dataset for different segmentation rates.}
\label{tab:maskPropagation}
\end{table}

\section{Conclusions}

DOT is a novel front-end algorithm for SLAM systems that robustly detects and tracks moving objects by combining instance segmentation and multi-view geometry equations. Our evaluation with ORB-SLAM2 in three public datasets for autonomous driving research \cite{KITTI}\cite{VKITTI}\cite{VKITTI2} demonstrates that DOT-generated object motion information allows us to segment the dynamic content, significantly improving its robustness and accuracy.

The independence of DOT from SLAM makes it a versatile front-end that can be adapted with minimal integration work to any state-of-art visual odometry or SLAM system. Unlike other systems, the mask tracking of DOT reduces the rate at which segmentation (typically involving high computational cost) should be done, reducing the computational needs with respect to the state of the art.

{\small
\bibliographystyle{ieee_fullname}
\bibliography{dot}
}

\end{document}